\documentclass[letterpaper]{article} 
\usepackage[hyphens]{url}  
\usepackage{times}
\usepackage{helvet}
\usepackage{courier}
\newcommand{\eg}{\emph{e.g.,}~}
\newcommand{\etal}{\emph{et al.,}~}
\newcommand{\ie}{\emph{i.e.,}~}
\usepackage{xcolor}
\usepackage{booktabs}
\usepackage{epsfig}

\usepackage{xcolor}
\usepackage{booktabs}
\usepackage{epsfig}
\usepackage{amsmath}
\usepackage{amssymb}
\usepackage{amsthm}

\usepackage{url}
\usepackage{tabularx}
\usepackage{multirow}
\usepackage{subfigure}
\usepackage{amsmath,dsfont}
\usepackage{array}
\usepackage{enumitem}
\usepackage{pifont}
\usepackage{threeparttable}

\usepackage{xcolor}
\usepackage{tikz-dependency}
\usepackage{pgfplots}
\usepackage{pgfplotstable}
\usepackage{makecell}
\usepackage{graphicx} 
\usepackage{natbib}  
\usepackage{caption} 
\usepackage{times}  
\usepackage{helvet}  
\usepackage{courier}  
\usepackage{algorithm}
\usepackage{algorithmic}

\usepackage[capitalize]{cleveref}
\Crefname{section}{Section}{Sections}
\Crefname{table}{Table}{Tables}
\crefname{figure}{Figure}{Figures.}

\usepackage{aaai24}

\newcommand{\mathbbm}[1]{\text{\usefont{U}{bbm}{m}{n}#1}} 

\DeclareMathOperator{\st}{s.t.}

\usepackage{newfloat}
\usepackage{listings}
\DeclareCaptionStyle{ruled}{labelfont=normalfont,labelsep=colon,strut=off} 
\floatstyle{ruled}
\newfloat{listing}{tb}{lst}{}
\floatname{listing}{Listing}
\title{CL2CM: Improving Cross-Lingual Cross-Modal Retrieval \\ via Cross-Lingual Knowledge Transfer}
\author{
    Yabing Wang\textsuperscript{\rm 1}\textsuperscript{\rm 2}\textsuperscript{\rm 3}\thanks{
    Work done during an internship at DAMO Academy.},
    Fan Wang\textsuperscript{\rm 3},
    Jianfeng Dong\textsuperscript{\rm 1}\textsuperscript{\rm 5}\thanks{Corresponding author},
    Hao Luo\textsuperscript{\rm 3}\textsuperscript{\rm 4}\textsuperscript{$\dagger$}
}
\affiliations{
    \textsuperscript{\rm 1} Zhejiang Gongshang University, \
    \textsuperscript{\rm 2} Xi'an Jiaotong University, \
    \textsuperscript{\rm 3}  DAMO Academy, Alibaba Group, \\
    \textsuperscript{\rm 4}  Hupan Lab, Zhejiang Province,
    \textsuperscript{\rm 5}  Zhejiang Key Lab of E-Commerce\\
    \{wyb7wyb7, dongjf24\}@gmail.com, \{Fan.w, michuan.lh\}@alibaba-inc.com

}

\usepackage{bibentry}

\begin{document}

\maketitle

\begin{abstract}

\begin{figure*}[tb!]
\centering\includegraphics[width=2.1\columnwidth]{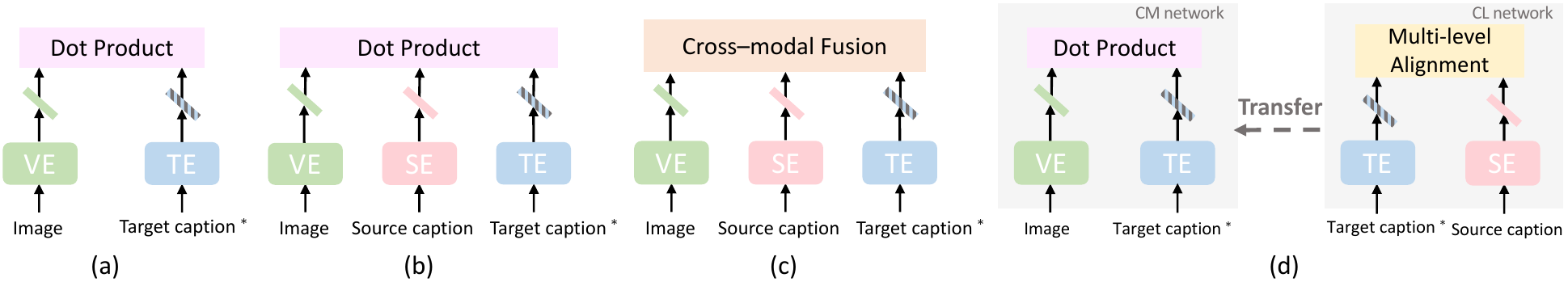}
\vspace{-2mm}
\caption{Comparison of different alignment methods for vision and target language: (a) Baseline method. (b) Universal CCR method, which projects the three data types into a common space by learning the alignment between them. (c) Single-stream method. (d) Proposed CL2CM method. Among these methods, (a), (b), and (d) are all dual-stream methods.
* indicates data with noise. VE'', SE'', and ``TE'' represent the image encoder, source-language encoder, and target-language encoder, respectively.
}\label{fig:title-pic}
\end{figure*}

Cross-lingual cross-modal retrieval has garnered increasing attention recently, which aims to achieve the alignment between vision and target language (V-T) without using any annotated V-T data pairs.
Current methods employ machine translation (MT) to construct pseudo-parallel data pairs,  which are then used to learn a multi-lingual and multi-modal embedding space that aligns visual and target-language representations. 
However, the large heterogeneous gap between vision and text, along with the noise present in target language translations, poses significant challenges in effectively aligning their representations.
To address these challenges, we propose a general framework, Cross-Lingual to Cross-Modal (CL2CM), which improves the alignment between vision and target language using cross-lingual transfer.
This approach allows us to fully leverage the merits of multi-lingual pre-trained models (\eg mBERT) and the benefits of the same modality structure, \ie smaller gap, to provide reliable and comprehensive semantic correspondence (knowledge) for the cross-modal network.
We evaluate our proposed approach on two multilingual image-text datasets, Multi30K and MSCOCO, and one video-text dataset, VATEX. The results clearly demonstrate the effectiveness of our proposed method and its high potential for large-scale retrieval.

\end{abstract}


\section{Introduction}\label{sec:introduction}

As the internet continues to expand globally, users from diverse linguistic backgrounds increasingly access multi-modal content online, such as images and videos.
Nevertheless, accurately retrieving such content poses considerable challenges, particularly for non-English speakers, as the majority of human-annotated datasets are only available in English. 
To address this issue, cross-lingual cross-modal retrieval (CCR) has emerged as a crucial area of research, aiming to develop models applicable to non-English languages without incurring substantial manual annotation costs.

The primary challenge for CCR lies in effectively achieving cross-lingual transfer and establishing reliable correlations between V-T.
A prevalent approach is to construct pseudo-parallel data pairs using MT and explicitly learn the correspondence between V-T.
As shown in \cref{fig:title-pic}(a), the straightforward baseline method directly attempts to align visual and target-language representations, similar to conventional cross-modal alignment methods \cite{kim2023exposing,pei2023clipping,fu2023learning,liu2022featinter,falcon2022feature,dong2022reading,shvetsova2022everything}.
But it suffers from the noisy translation issue.
Previous efforts \cite{wang2022cross,jain2021mural,zhang2022generalizing} attempt to alleviate this issue by incorporating the matching between visual and source-language representations, and treat the source language as the focal point, as shown in \cref{fig:title-pic}(b).
However, these methods are limited to instance-level matching \footnote{Instance level matching refers to the matching between individual image and text instances, and the alignment loss is applied to the final embedding layer, similar to the CLIP model.} in the visual-text domain, as they match image-text pairs relying on global feature vectors.
On the other hand, some endeavors \cite{huang2021multilingual,ni2021m3p,zhou2021uc2,cclm} utilize cross-modal fusion modules to model fine-grained interactions between image regions and target-language words, as depicted in \cref{fig:title-pic}(c). Although these methods demonstrate promising performance, they entail huge computational costs, as all possible query-candidate pairs need to be fed into the fusion modules during inference. 
In light of the above discussions, we argue that conventional approaches still face \textit{a dilemma between achieving reliable alignment of V-T and computational efficiency.} 

In this paper, we propose a novel solution: \textbf{improving the alignment between vision and target language using cross-lingual knowledge transfer}.
We note that existing multi-lingual pre-trained models, such as mBERT \cite{devlin2018bert}, have demonstrated remarkable performance in cross-lingual alignment but have not been fully utilized in CCR tasks, where they are typically only used as multi-lingual text backbones.
Furthermore, cross-lingual sentences exhibit more analogous modality structures, with a smaller gap compared to visual elements.
These properties provide a unique opportunity to comprehensively explore the knowledge between different languages and transfer it to align V-T.

In specific, we propose a novel framework named \underline{C}ross-\underline{L}ingual to \underline{C}ross-\underline{M}odal (CL2CM) that aims to address two main challenges mentioned above: 1) noisy translations and 2) reliable and comprehensive correspondence learning. 
As shown in \cref{fig:title-pic}(d), CL2CM consists of a cross-lingual (CL) network with multi-level alignment and a cross-modal (CM) network for vision-target language matching. 
The multi-level alignment in the CL network is designed to tackle the noisy translation issue, including both sentence-level and self-supervised word-level alignment.
For self-supervised word-level alignment, we generate pseudo-labels in a self-supervised manner to align cross-lingual representations. This approach mitigates misalignment caused by noisy words, and enhances the discrimination ability of the representations at the word level.
To address the second challenge,
we employ the knowledge distillation to transfer the CL knowledge to the CM network. The CL network, utilizing multi-level alignment, can provide more comprehensive and reliable correspondence, which can also alleviate the effect of noisy translations (\ie noise correspondence \cite{huang2021learning}).
\textit{Moreover, during inference, only the cross-modal network is utilized in CL2CM, bringing no additional cost compared to other dual-stream methods.}

To sum up, our contributions can be summarized as follows: 
(i) We propose a novel framework called CL2CM that improves the alignment between vision and target language using cross-lingual knowledge transfer. To the best of our knowledge, we are the first to explore cross-lingual transfer for CCR.
(ii) We introduce a multi-level alignment strategy to explore the cross-lingual knowledge and alleviate the effect of noisy translations.
(iii) Extensively experiments on three CCR benchmarks, \ie Multi30K, MSCOCO, and VATEX, demonstrating the effectiveness of our proposed method and its high potential for large-scale retrieval.


\section{Related Work} \label{sec:rel-work}

\subsection{Cross-lingual Cross-modal Retrieval}
Cross-lingual cross-modal retrieval has been garnering increased attention amongst researchers, as it enables the acquisition of images or videos utilizing a non-English query, without relying on human-labeled vision-target language data \cite{li2023svitt,wu2023cap4video,li2022align,wei2021universal}. This method mitigates the constraints of conventional cross-modal retrieval tasks \cite{yang2020tree,dong2022partially,zheng2023progressive,hu2022lightweight} centered on English, and offers a highly efficient and cost-effective solution for target-language based retrieval, greatly reducing the need for human-labeled data.
In terms of the model architecture, there are mainly two broad directions to conduct CCR.
The first line \cite{zhou2021uc2,ni2021m3p,cclm} utilizes a single-stream model that incorporates cross-lingual and cross-modal fusion modules to model both image regions and multilingual text word representations in a unified semantic space, capturing the fine-grained relationship between them. 
For example, Zhou \etal \cite{zhou2021uc2} use a cross-lingual cross-modal encoder to model the interaction between vision and multiple languages, employing the masked modeling loss.
However, this approach has limitations in real-world scenarios, such as high computational costs and reliance on object detectors for image regions, rendering it less practical for large-scale CCR tasks.

The other line \cite{wang2022cross,huang2021multilingual,jain2021mural,zhang2022generalizing,wang2023dual} involves two-stream models, where each stream is dedicated to modeling the vision or language inputs.
For example, Jain et al. \cite{jain2021mural} learn V-T alignment by combining image-text matching and text-text matching tasks, using scalable dual encoder models trained with contrastive losses.
Although these approaches are efficient, they only consider the instance-level alignment but ignore fine-grained correspondence between visual and textual information, resulting in suboptimal performance. 
In our work, we dedicate the two-stream structure, and leverage cross-lingual knowledge to improve the alignment quality between vision and the target language.

\subsection{Knowledge Distillation}
Knowledge Distillation is defined as a learning manner that  extracts ``dark knowledge" from a teacher network to guide the learning of a student network, encouraging the student model to imitate the teacher’s behavior. 
This technique is increasingly important for model compression and transfer learning.
Hinton et al. \cite{hinton2015distilling} first proposed minimizing the KL-divergence of category distributions for classification tasks by matching logits produced by two models.
Except for final layer logits distillation, some methods \cite{romero2015fitnets,sun2020mobilebert} use knowledge distillation to distill compact feature representations from the teacher network, such as, Sun \etal \cite{sun2020mobilebert} proposed MobileBERT to compress and accelerate the popular BERT model.

Moreover, in the multi-modal field, knowledge distillation can be applied to compress visual-and-language (VL) models \cite{fang2021compressing,rao2021student,dong2023dual}. For instance, Fang \etal \cite{fang2021compressing} employ knowledge distillation to effectively compress a transformer-based large VL model into a small VL model.
In our work, we use knowledge distillation to transfer the CL knowledge to improve the alignment between vision and target language.


\section{Methods}
\label{sec:method}

\begin{figure*}[tb!]
\centering\includegraphics[width=2\columnwidth]{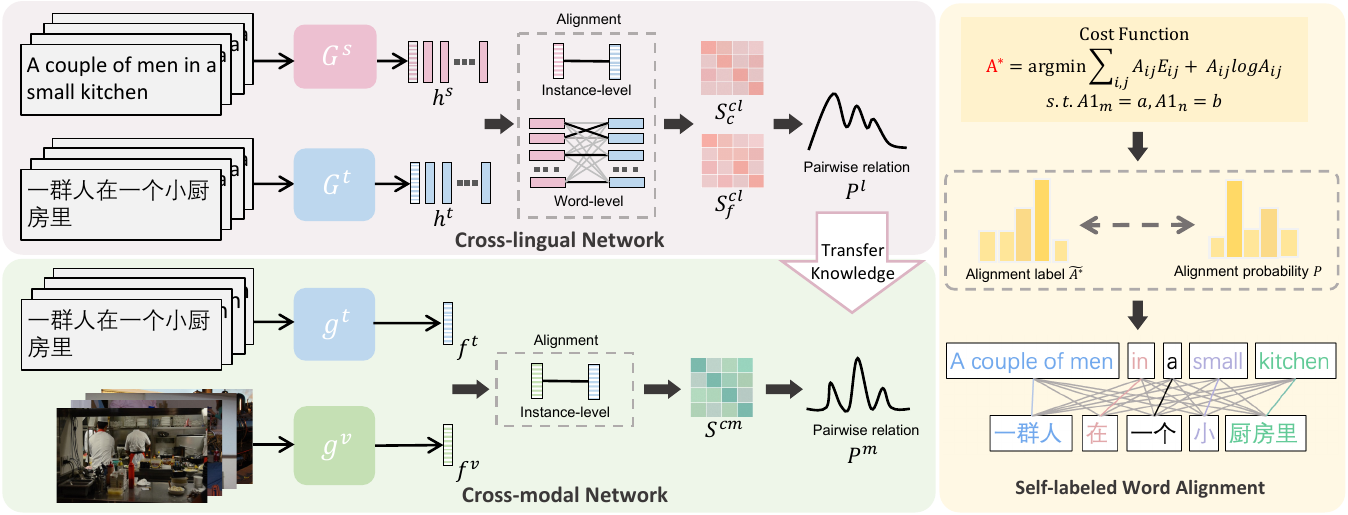}
\vspace{-2mm}
\caption{Overview of the proposed CL2CM framework. It consists of a cross-lingual (CL) with multi-level alignment and a cross-modal (CM) network with instance-level alignment. We aim to improve the alignment quality between vision and target-language using cross-lingual knowledge transfer.
}\label{fig:framework}
 \vspace{-3mm}
\end{figure*}

\subsection{Preliminary}
We consider a dataset $\mathcal{D} = {(V, S)}$ consisting of annotated paired images (or videos) $V$ with source-language captions $S$. 
However, obtaining human-labeled V-T pairs during training may not always be feasible. 
To address this issue, we follow the approach in \cite{wang2022cross} and create an extended dataset $\hat{\mathcal{D}} = {(V, S, T) }$, where $T$ indicates the translated target-language captions corresponding to $S$. Note that during inference, we use only human-input target-language sentences as queries for retrieval. 

\cref{fig:framework} presents an overview of our approach.
In what follows, we first depict the CL network with multi-level alignment, followed by the description of the CM network, and finally, introduce the CL knowledge transfer.
For ease of description, we use image-text retrieval as an example to describe our framework.

\subsection{Cross-lingual Network}\label{sec:sec1}

Given a source-language caption $s$ consisting of $M$ words and its corresponding target-language caption $t$ consisting of  $N$ words, we utilize a pre-trained multilingual encoder $G(\cdot)$ to extract the contextualized word embeddings $h^s = \{\overline{h^s}; h^s_{1}, ..., h^s_{M}\}$ and $h^t = \{\overline{h^t}; h^t_{1}, ..., h^t_{N}\}$, respectively. Here, $\overline{h^s}$ and $\overline{h^t}$ are [CLS] tokens, and we take them as sentence-level representations. It can be defined as: 
\begin{equation}
\begin{array}{cc}
h^s = G^s(s) ,
h^t = G^t(t)
\end{array}
\end{equation}

To capture the complex correspondence and align the presentations between cross-lingual data pairs accurately, we develop a multi-level alignment strategy that establishes correlations between captions in different languages at various levels of granularity.

\textbf{Instance-level alignment.} 
Given a batch of $B$ pseudo-parallel sentence pairs, 
we first compute the instance-level similarity scores $\mathcal{S}^{ins}$ using sentence-level representations. Then, we utilize a symmetric InfoNCE loss over the similarity matrix to optimize the text encoder and learn the discriminative cross-lingual features. It can be formulated as:

\begin{equation}
\begin{array}{cc}
\mathcal{L}^{cl}_{instance} = - \frac{1}{2} \times \frac{1}{B} \sum^B_{i=1} [ log \frac{exp(\mathcal{S}^{ins}_{i,i})}{\sum^B_{j=1}exp(\mathcal{S}^{ins}_{j,i})} \vspace{1.5ex}\\
+ log \frac{exp(\mathcal{S}^{ins}_{i,i})}{\sum^B_{j=1}exp(\mathcal{S}^{ins}_{i,j})} ] 
\end{array}
\end{equation}
where $\mathcal{S}^{ins}_{i,j} = \frac{{\overline{h^s}_i}^T \overline{h^t}_j}{||\overline{h^s}_i|| \cdot ||\overline{h^t}_j||} $, and $\overline{h^s}_i, \overline{h^t}_j \in \mathbb{R}^d$ represent the $i$-th  source-language and $j$-th target-language sentence representations, respectively.
The objective is to learn more distinctive cross-lingual sentence representations.

\textbf{Self-supervised word-level alignment.} 
Although instance-level alignment can promote cross-lingual matching and learn discriminative sentence features, 
it may cause the text encoder to only focus on abstract semantic information and lose its original word-level discriminative power.
However, the word-level label is inaccessible, and while some word alignment tools are available, they may not be suitable for all languages, particularly for low-resource languages.
To overcome this issue, we propose a self-supervised word alignment method that does not require additional word alignment tools to align cross-lingual captions at the word level. 
Specifically, we define the objective as computing an alignment matrix $A$ between source and target language words, and then maximizing the probability distribution $P$ of each aligned word pair. This optimization problem can be formulated as:

\begin{equation}
\label{eq:word_alignment}
    \underset{A}{\min} \sum_{m,n} -A_{m,n}logP(s_m, t_n)
\end{equation}
 
To solve this equation, the Optimal Transport (OT) theory can be incorporated into our word alignment approach.
OT has been applied in domain adaptation \cite{xu2020reliable,chang2022unified} to align
the representations in the source and target domains with
associated theoretical guarantees.
It encourages a global mapping to mine domain statistics property for discovering intrinsic differences among clean and noisy sample pairs.
We treat the source and target sentences as two probability distributions and employ OT to find the optimal alignment between them.
This can be formally defined as:
\begin{equation}
\begin{array}{cc}
    \underset{A}{\min} \sum_{m,n} A_{m,n}C(s_m,t_n) \\
    \st \ A \mathbbm{1}_{M} = \frac{1}{N} \mathbbm{1}_{N}, \ A^{T} \mathbbm{1}_{N} = \frac{1}{M}\mathbbm{1}_{M}
\end{array}
\end{equation}
where $\mathbbm{1}_{D}$ represents a D-dimensional vector whose elements are all 1. 
The cost matrix $C(s_m,t_n) \in \mathbb{R}^{m \times n}$ represents the cost of aligning the $m$-th word in $S$ with the $n$-th word in $T$, which can be formulated as:
\begin{equation}
\begin{array}{cc}
    C(s_m,t_n) =  \frac{{h^s_m}^T h^t_n}{||h^s_m|| \cdot ||h^t_n||} \vspace{0.5ex}\\
\end{array}
\end{equation} 
To approximate OT efficiently, we add an entropic regularizer $E(A)$ to the optimization problem, as shown below: 
\begin{equation}
\begin{array}{cc}
\label{eq:optimal_transport_entropy}
    \underset{A}{\min} \sum_{m,n} A_{m,n}C(s_m,t_n) + \lambda E(A) \vspace{0.5ex}\\
    \st \ A \mathbbm{1}_{M} = \frac{1}{n}\mathbbm{1}_{N}, \ A^{T} \mathbbm{1}_{N} = \frac{1}{M}\mathbbm{1}_{M}
\end{array}
\end{equation}
where $E(A) = \mu A logA$. The \cref{eq:optimal_transport_entropy} has a unique solution $A^*$ such that: 
\begin{equation}
\begin{array}{cc}
    A^* = diag(\mu) K diag(v) \vspace{1ex}  \\
    K_{m,n} = e^{C(s_m, t_n) / \mu}
\end{array}
\end{equation}
where $\mu \in \mathbb{R}^{M}_+$, $v \in \mathbb{R}^{N}_+$, $K \in \mathbb{R}^{N \times M}_+ $, which solved with Sinkhorn’s fixed point iteration:
\begin{equation}
\begin{array}{cc}
u^{(t+1)} = \frac{\mathbbm{1}_N}{Kv^{(t)}}, \ v^{(t+1)} =  \frac{\mathbbm{1}_M}{K^T u^{(t+1)}}
\end{array}
\end{equation}

With the solved stochastic matrix $A^*$, we can produce the alignment labels $\widetilde{A^*}$ by applying a threshold $\gamma$ and a function $h(,)$ to $A^*$:
\begin{equation}
\begin{array}{cc}
\widetilde{A^*}_{m,n} = \frac{h(A^*_{m,n}, \gamma) \times A^*_{m,n}}{\sum_{n=1}^{N} h(A^*_{m,n}, \gamma)} \vspace{1ex} \\
 h(A^*_{m,n}, \gamma) = 1 - \frac{1}{2}(1+sgn(A^*_{m,n} - \gamma)) \vspace{0.5ex}\\
\end{array}
\end{equation}
where $\gamma$ is a threshold of alignment, which we set to the mean value of A$^{*}$, and $sgn(x)=1$ when $x>0$, and equals $-1$ otherwise. The training objective is to maximize the alignment probabilities between aligned words. This can be defined as:
\begin{equation}
\begin{array}{cc}
\mathcal{L}_{word} = \sum \limits _{m,n}- \widetilde{A^*}_{m,n} logP(s_m, t_n) \vspace{1ex} \\
P(s_m, t_n) =  \frac{ exp( C(s_m, t_n))}{\sum_{n=1}^{n} exp( C(s_m, t_n))}
\end{array}
\end{equation}

\begin{figure}[tb!]
\centering\includegraphics[width=0.9\columnwidth]{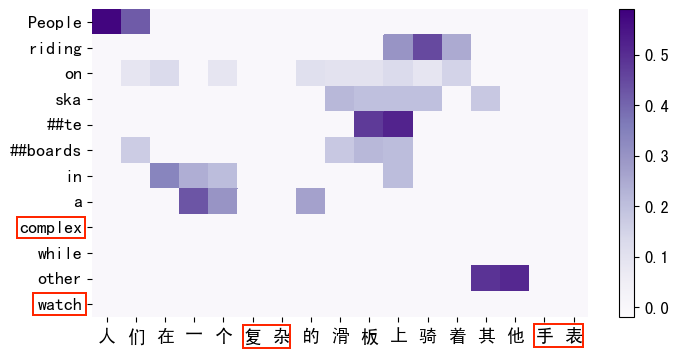}
\vspace{-2mm}
\caption{
Visualization of the generated pseudo-label in self-supervised word-level alignment. The red box represents the incorrect translated word of the corresponding source-language word.
}\label{fig:transfer-knowledge-visual}
\end{figure}

The objective function described above is designed to minimize the distance between the best-matched word pairs while increasing the distance between poorly matched pairs. This helps alleviate the problem of misalignment caused by noisy words to some extent, as shown in \cref{fig:transfer-knowledge-visual}. 
Finally, we can align words in different languages and identify the best matching result from a candidate word set.
To train the cross-lingual network, we define the objective function as $\mathcal{L}^{cl} = \mathcal{L}^{cl}_{instance} + \mathcal{L}_{word}$.
Besides, although other OT variants have existed (see \cref{tab:ot}), this does not affect the starting point of this paper.

\subsection{Cross-modal Network}
\label{sec:cross-modal-network}

The cross-modal network comprises an image encoder $g^v(\cdot)$ and a target-language encoder $g^t(\cdot)$, which can produce an image feature vector and a target-language caption feature vector, respectively. We denote them as $f^v$ and $f^t \in \mathbb{R}^d$, respectively. To enhance the target-language representation, we share $g^t(\cdot)$ with $G^t(\cdot)$ in the CL network.
\begin{equation}
\begin{array}{cc}
f^v = g^v(V) \\
f^{t} = g^{t}(T)
\end{array}
\end{equation}
The goal of the cross-modal network is to accurately learn the alignment between visual and target-language representations. To achieve this, we define the objective function as:
\begin{equation}
\begin{array}{cc}
\mathcal{L}^{cm} = \alpha \cdot \mathcal{L}^{cm}_{instance} + (1-\alpha) \cdot \mathcal{L}_{ckt} 
\end{array}
\end{equation}
where $\alpha$ is a weight hyper-parameter, $\mathcal{L}^{cm}_{instance}$ is the instance-level alignment objective, which aligns image and target-language features at the instance level using the global vectors, and $\mathcal{L}_{ckt}$ represents the cross-lingual knowledge transfer loss which will be introduced in next Section. 
The $\mathcal{L}^{cm}_{instance}$ can be computed as:
\begin{equation}
\begin{array}{cc}
\mathcal{L}^{cm}_{instance} = \frac{1}{2} \times \frac{1}{B} \sum^B_{i=1} [ log \frac{exp(\mathcal{S}^{cm}_{i,i})}{\sum^B_{j=1}exp(\mathcal{S}^{cm}_{j,i}))} \vspace{1.5ex}\\
+ log \frac{exp(\mathcal{S}^{cm}_{i,i})}{\sum^B_{j=1}exp(\mathcal{S}^{cm}_{i,j})} ] 
\end{array}
\end{equation}
where $\mathcal{S}^{cm}_{i,j} = \frac{{f^v_i}^T f^t_j}{||f^v_i|| \cdot ||f^t_j||} $ represents the instance-level similarity of $i$-th image and $j$-th target language caption.

The final training objective can be formulated as $\mathcal{L} = \mathcal{L}^{cm} + \mathcal{L}^{cl}$. It's worth noting that we only use the CM network to calculate the instance-level similarity matrix $\mathcal{S}^{cm}$ for image-target language retrieval during inference.

\begin{figure}[tb!]
\centering\includegraphics[width=1.0\columnwidth]{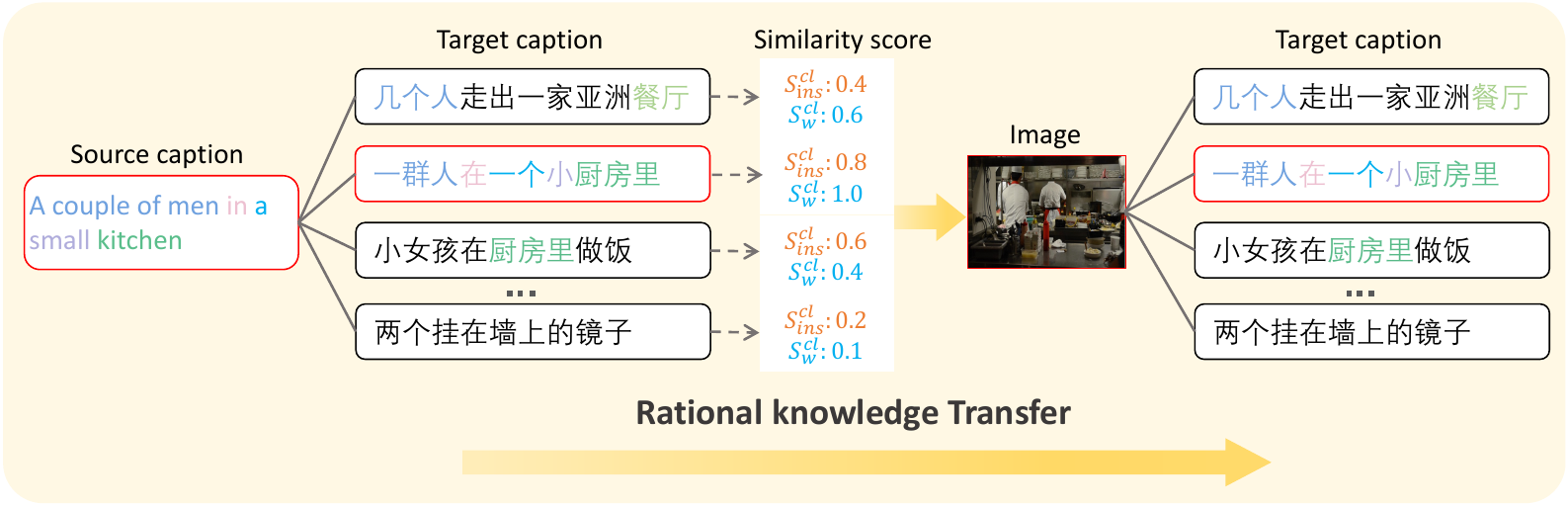}
\vspace{-7mm}
\caption{
An illustration of the CL knowledge transfer. 
}\label{fig:transfer-knowledge}
\end{figure}

\subsection{Cross-lingual Knowledge Transfer}
\label{sec:rkt}

The significant heterogeneity gap between images and target languages may pose challenges in establishing their correlations. However, the reliable semantic relations are crucial for cross-modal retrieval, as the essence of retrieval is ranking.
Considering the CL network uses a multi-lingual pre-trained model and training with multi-level alignment, enabling it to model more reliable and comprehensive correspondence.
Therefore, we introduce Relational Knowledge Distillation \cite{park2019relational} to transfer the relational knowledge learned by the CL network to the CM network to provide the extra discriminative information, as shown in \cref{fig:transfer-knowledge}.
Specifically, we propose a multi-level cross-lingual knowledge transfer that involves a voting ensemble of two-level structural relational similarities (i.e., sentence and word levels). A high similarity score at both levels indicates a strong correlation between the sample pairs. The CL knowledge can be formulated as follows:

\begin{equation}
\begin{array}{cc}
\mathcal{S}^{cl}(s_i, t_j) = \lambda \cdot \mathcal{S}^{cl}_{sent}(s_i, t_j) + (1 - \lambda) \cdot \mathcal{S}^{cl}_{word}(s_i, t_j) \vspace{1ex} \\
\mathcal{S}^{cl}_{word}(s_i, t_j) = \frac{1}{M} \sum \limits_{m=1}^M  \underset{0 \leq n \leq N}{\max}   \frac{{h^s_m}^T h^t_n}{||h^s_m|| \cdot ||h^t_n||} \vspace{1ex} 

\end{array}
\end{equation}
where $\lambda$ is a weight hyper-parameter, and $\mathcal{S}^l_{word}$ is calculated from the word-level cross-lingual representations that can capture the subtle semantic difference between data pairs.
By combining the two-level similarities, CL knowledge accurately depicts mutual relations beyond what each individual level can achieve alone. The CL knowledge transfer loss is then formulated as:
\begin{equation}
\begin{array}{cc}
p^{cl} = softmax(\mathcal{S}^{cl} / \tau) \vspace{1ex}\\
p^{cm} = softmax(\mathcal{S}^{cm} / \tau)\\ 
\mathcal{L}_{rkt} = \frac{1}{B} \sum \limits^B_{i=1} \sum KL(p^{cm}_i \ || \ p^{cl}_i)
\end{array}
\end{equation}
where $KL$ denotes the Kullback-Leiber divergence, and $\tau$ is a temperature parameter. 
By transferring the CL knowledge to the CM network, we can improve the quality of alignment between visual and target-language representations, ultimately enhancing the retrieval performance.

\section{Experiment} \label{sec:eval}

\subsection{Experimental Settings} \label{ssec:exp-set}

\textbf{Datasets.}
We conduct experiments on two public multilingual image-text retrieval datasets (Multi30K and MSCOCO), as well as a video-text retrieval dataset (VATEX). Notably, we only use the annotated vision-source language data pair in the training process, while using the annotated vision-target language data pairs during inference.

\begin{itemize}[leftmargin=*]
\item \textbf{Multi30K}~\cite{elliott2016multi30k}: 
This dataset consists of 31,000 images and is a multi-lingual version of Flickr30K \cite{young2014image}. 
It involves four languages, \ie English(en), German(de), French(fr), and Czech(cs).
We adopt a similar data partition as \cite{young2014image}.
\item \textbf{MSCOCO} \cite{chen2015microsoft}: This dataset consists of 123,287 images, and each image has 5 captions. We translate the training set from English into Chinese(zh) and Japanese(ja) by resorting to MT, and using the test sets from the \cite{li2019coco} and \cite{yoshikawa2017stair}, respectively. We follow the data split as in \cite{zhou2021uc2}.
\item \textbf{VATEX}~\cite{wang2019vatex}:
This is a bilingual video-text retrieval dataset with over 41,250 videos, each paired with 10 English and 10 Chinese sentences. We use only the annotated English captions from the training set and generate the corresponding Chinese translations using MT. We adopt a similar data partition as \cite{chen2020fine}.

\end{itemize}

\textbf{Evaluation metrics.}
For video-text retrieval, we follow the previous works~\cite{cvpr2019-dual-dong,li2019w2vv++}, and use rank-based metrics, namely $R@K$($K = 1, 5, 10$), mean Average Precision (mAP), and sum of all Recalls (SumR) to evaluate the performance. $R@K$ is the fraction of queries that correctly retrieve desired items in the top $K$ of ranking list.
Following previous works \cite{zhou2021uc2,cclm}, we only report the SumR for image-text retrieval.

\textbf{Implementation Details.}
\label{sec:implementation_details}
We apply the CLIP (ViT-B/32) \cite{radford2021learning} and mBERT-base \cite{devlin2018bert} as the image and text encoder, respectively.
For video encoder, we adopt I3D \cite{carreira2017quo} video features and use multi-layer perceptron followed by mean-pooling. 
Besides, we set $\lambda=0.6$, $\alpha=0.4$, and $\tau=0.07$ in our experiments. The batch size is 128, and an Adam optimizer with an initial learning rate 2.5e-5 and adjustment schedule similar to \cite{luo2022clip4clip} is utilized.

\subsection{Ablation Studies}

\begin{table}[tb!]
\setlength{\abovecaptionskip}{2mm} 
\setlength{\belowcaptionskip}{0cm}
\caption{Ablation study of each model component on Multi30K. ``MLA" and ``CLKT" indicate the multi-level alignment and CL knowledge transfer, respectively.
}
\label{tab:component-ablation}
\centering 
\scalebox{0.8}{
\begin{tabular}{@{}l*{12}{r}c @{}}
\toprule
\multicolumn{2}{c}{\textbf{MLA}} &&\multicolumn{2}{c}{\textbf{CLKT}} & \multirow{2}{*}{\textbf{en2de}} & \multirow{2}{*}{\textbf{en2fr}} &\multirow{2}{*}{ \textbf{en2cs}} \\
\cmidrule{1-2} \cmidrule{4-5}
\textbf{$L^{cl}_{instance}$} & \textbf{$L_{word}$} && \textbf{$S^{cl}_{sent}$}  & \textbf{$S^{cl}_{word}$}   \\ 
\hline
&       &      && &473.2 &482.3  &472.6\\
\makecell{\ding{51}}  &  &&  &  & 488.1  & 488.7 &476.0       \\
\makecell{\ding{51}}  & \makecell{\ding{51}}  &&  & &490.0  &493.0  & 480.4        \\
\makecell{\ding{51}}  &   && \makecell{\ding{51}}  &    &491.5   &   492.9  & 480.7      \\
\makecell{\ding{51}}  & \makecell{\ding{51}}  && \makecell{\ding{51}}  &                                                        &493.0       &494.4       &481.2       \\
\makecell{\ding{51}}                                                          & \makecell{\ding{51}} &&  & \makecell{\ding{51}}  & 494.3      & 496.1       & 483.6      \\
\makecell{\ding{51}}  & \makecell{\ding{51}}  && \makecell{\ding{51}}  & \makecell{\ding{51}}  &498.0 &499.7   &485.8       \\
\bottomrule
\end{tabular}
}
\end{table}

\textbf{Impact of the component modules.}
We analyze the effect of each component using the control variable method in \cref{tab:component-ablation}. The first line shows the performance of the Baseline method without the CL network.
Based on the Baseline, we further add multi-level alignment and CL knowledge transfer. Both methods result in significant improvements, demonstrating the effectiveness of our approach in enhancing the alignment between V-T from two perspectives.
Moreover, we observe a steady improvement when gradually adding our two components. This means that our two ideas are complementary and effective.

\noindent\textbf{The exploration of self-supervised word-level alignment. }
To further investigate the generation method of pseudo-label, we attempt to replace the OT-based method with the cross-attention module. In specific, we use source-language word representations as query, target-language ones as key and value, and apply generated attention as the pseudo-label. As shown in \cref{tab:word-alignment}, the OT-based method yields significantly better results than the cross-attention counterpart.
We speculate that the reason for this difference is that the OT-based method is more robust and better able to handle noisy translations than cross-attention.

\begin{table} [tb!]
\renewcommand{\arraystretch}{1.1}
\setlength{\abovecaptionskip}{2mm} 
\setlength{\belowcaptionskip}{0cm}
\caption{
Ablation study of generating pseudo-labels in self-supervised word-level alignment on Multi30K. ``Cross-attention" indicate that generating pseudo word alignment label relies on the cross-attention module.}
\label{tab:word-alignment}
\centering 
\scalebox{0.9}{
\begin{tabular}{@{}l*{12}{r}c @{}}
\toprule
\textbf{Method} & {\textbf{en2de}} & {\textbf{en2fr}}  & {\textbf{en2cs}}\\
\hline
Cross-attention &{495.7} &{496.1} &{480.4}\\
Ours &\textbf{498.0} &\textbf{499.7} &\textbf{485.3} \\
\bottomrule
\end{tabular}
}
\end{table}

\begin{figure}[tb!]
\centering
\subfigure[Two-stage CL2CM variants training pipeline]{
\label{variants-a}
\includegraphics[width=0.8\columnwidth]{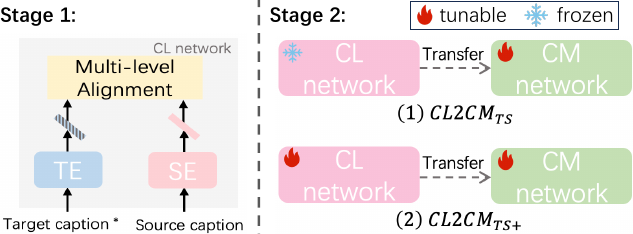}
}
\subfigure[Performance comparison of different variants]{
\label{variants-b}
\includegraphics[width=0.8\columnwidth, height=100px]{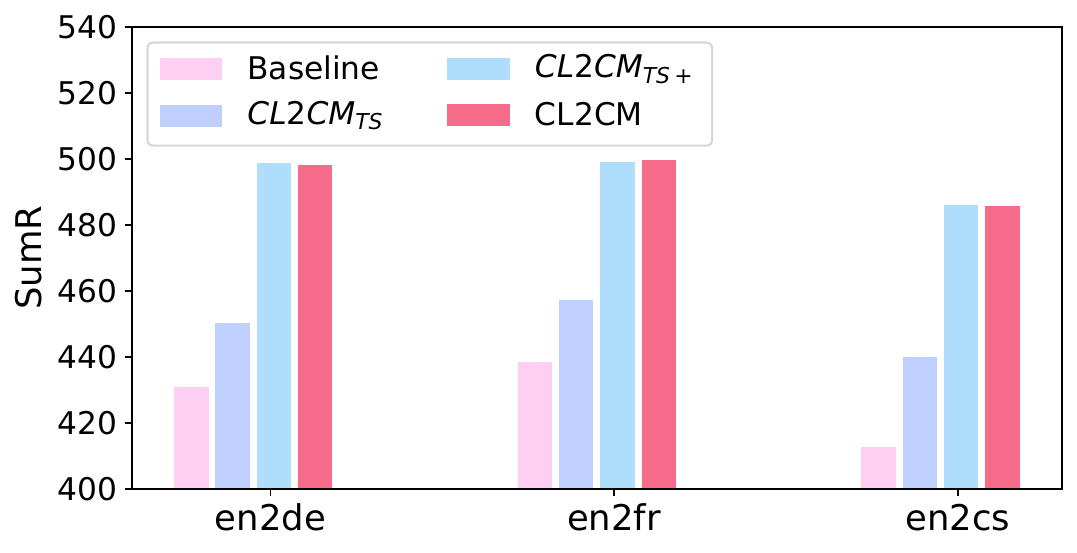}
}
\vspace{-4mm}
\caption{
Ablation study to investigate the impact of different CL knowledge transfer approaches on Multi30K. Two-stage CL2CM variants initially train a CL network, followed by knowledge transfer to the CM network. The results indicate that CL knowledge transfer achieves significant performance gains, and the multi-level alignment in stage 2 can further improve the target-language representation learning.
}\label{fig:transfer-knowledge-fashion}
\end{figure}

\noindent\textbf{The exploration of different model variants. }
\cref{fig:transfer-knowledge-fashion} illustrates the performance of different model variants. 
We find that incorporating CL knowledge transfer led to better performance compared to the Baseline method. 
However, the CL2CM$_{TS}$ method, which only provides CL knowledge without representation improvement, resulted in sub-optimal results. Additionally, the use of the two-stage training did not lead to significant performance gains. Therefore, we used the end-to-end training strategy in subsequent experiments. 
We suspect that the pre-trained language models used in our experiments already have strong cross-lingual alignment capabilities, and the dataset we used is relatively small. Moreover, even without the first-stage training, cross-lingual alignment can still achieve rapid convergence. While using additional parallel corpora in the first stage may lead to further performance improvements, our goal is to demonstrate the effectiveness of cross-lingual transfer. We leave further exploration in this direction to future research.

\noindent\textbf{The exploration of different OT variants. }
\begin{table} [tb!]
\renewcommand{\arraystretch}{1.1}
\setlength{\abovecaptionskip}{1.5mm} 
\setlength{\belowcaptionskip}{-0.05cm}
\caption{
Performance of our proposed model with varied OT variants on Multi30K.
}
\label{tab:ot}
\centering 
\scalebox{0.9}{
\begin{tabular}{@{}l*{12}{r}c @{}}
\toprule
\textbf{Method} & {\textbf{en2de}} & {\textbf{en2fr}}  & {\textbf{en2cs}}\\
\hline
Unbalance OT &{497.2} &{496.8} &{484.0}\\
Partial OT &{497.3} &{497.1} &{484.5}\\
Vanilla OT &{498.0} &{499.7} &{485.3} \\
\bottomrule
\end{tabular}
}
\end{table}
In \cref{tab:ot}, we experimented with varied OT variants (\ie unbalanced OT, partial OT, and vanilla OT). The results show that the performance did not change significantly, demonstrating that our approach is compatible with other OT methods. 
Furthermore, these OT variants all aim to find an effective optimal matching solution, although the implementation may differ, this does not affect the core contribution of this paper. In our experiment, we apply the vanilla optimal transport.

\begin{table} [tb!]
\renewcommand{\arraystretch}{1.0}
\setlength{\abovecaptionskip}{1.5mm} 
\setlength{\belowcaptionskip}{-0.05cm}
\caption{
Cross-lingual image-text retrieval results on Multi30K and MSCOCO (the source language is English and the target language is non-English). 
*: Models pre-trained on large-scale datasets, e.g., CC3M and its MT version. \dag : Model uses the same initialization parameters in the backbone with CCLM. The metric is the sum of all Recalls (sumR) following previous works.
The single-stream method is usually a one-to-one matching Siamese architecture, so its inference efficiency is lower than that of the two-stream method. During inference, CL2CM$^\dag$ is 11x faster than CCLM.
}
\label{tab:sota-multi30k-and-mscoco}
\centering 
\scalebox{0.7}{
\begin{tabular}{@{}l*{12}{r}c @{}}
\toprule
\multirow{2}{*}{\textbf{Method}}  & \multicolumn{3}{c}{\textbf{Multi30K}} && \multicolumn{2}{c}{\textbf{MSCOCO}}\\
\cmidrule{2-4} \cmidrule{6-7}
& {\textbf{en2de}} & {\textbf{en2fr}}  & {\textbf{en2cs}} && {\textbf{en2zh}}  & {\textbf{en2ja}}\\
\hline
\textbf{Single-Stream:} \\
M$^3$P \cite{ni2021m3p}* &351.0 &276.0 &220.8 &&332.8 &336.0\\
UC$^2$ \cite{zhou2021uc2}*  &449.4 &444.0 &407.4 &&492.0 &430.2\\
CCLM  \cite{cclm}*  &\textbf{540.0} &\textbf{545.4} &\textbf{536.4} &&\textbf{546.0} &\textbf{532.8}\\
\hline
\textbf{Two-Stream:} \\
MURAL \cite{jain2021mural}*  &456.0 &454.2 &409.2 &&- &435.0\\
MLA \cite{zhang2022generalizing}*  &495.6 &510.0 &457.2 &&- &482.4\\
NRCCR \cite{wang2022cross} &{480.6} &{482.1} &{467.1} &&{512.4} &{507.0}\\
CL2CM  &{498.0} &{499.7} &{485.3} && {522.0} & {515.9}\\
CL2CM$^\dag$ &\textbf{530.4} &\textbf{536.0} &\textbf{526.3} &&\textbf{544.3} &\textbf{546.2}\\
\bottomrule
\end{tabular}
 }
\end{table}

\begin{table*} [tb!]
\renewcommand{\arraystretch}{1.2}
\setlength{\abovecaptionskip}{1.5mm} 
\setlength{\belowcaptionskip}{0cm}
\caption{
Cross-lingual video-text retrieval results on VATEX (the source language is English and the target language is Chinese). 
*: Model pre-trained on a large-scale dataset Multi-HowTo100M \cite{huang2021multilingual}.
}
\label{tab:sota-vatex}
\centering 
\scalebox{1.0}{
\begin{tabular}{@{}l*{12}{r}c @{}}
\toprule
\multirow{2}{*}{\textbf{Method}}   & \multicolumn{4}{c}{\textbf{Text-to-Video Retrieval}} && \multicolumn{4}{c}{\textbf{Video-to-Text Retrieval}} & \multirow{2}{*}{\textbf{SumR}} \\
 \cmidrule{2-5}  \cmidrule{7-10} 
& R@1 & R@5 & R@10  & mAP && R@1 & R@5 & R@10  & mAP & \\
\cmidrule{1-11}
MMP \cite{huang2021multilingual}  &23.9 &55.1 &67.8 &-  && - &- &- &-  &-\\
MMP \cite{huang2021multilingual}* &29.7 &63.2 &{75.5} &-  && - &- &-  &- &-  \\
{NRCCR} \cite{wang2022cross}  & {30.4}  & {65.0}  &{75.1}    &{45.64} && {40.6}  &{72.7}  &{80.9}  &{32.40} & {364.8} \\
{CL2CM} & \textbf{32.1}  & \textbf{66.7}  &\textbf{77.3}   &\textbf{47.49} && \textbf{48.2}  
&\textbf{77.1}  &\textbf{85.5}    &\textbf{35.77} & \textbf{386.9} \\
[3pt]
\bottomrule
\end{tabular}
 }
\end{table*}

\subsection{Evaluation on Cross-lingual Image-Text Retrieval}

We evaluate the performance of our CL2CM method against state-of-the-art approaches on two widely used image-text retrieval datasets, \ie Multi30K and MSCOCO. 
Notably, M$^3$P, UC$^{2}$, CCLM, MURAL, and MLA were pre-trained on large-scale multi-lingual vision-language datasets, while NRCCR and our method do not require additional pre-training data.
As shown in \cref{tab:sota-multi30k-and-mscoco}, our CL2CM outperforms all two-stream methods, achieving significant performance gains. 
When equipped with a powerful backbone (\ie SwinTransformer for image encoding and XLM-R for text encoding), CL2CM$^\dag$ achieved significant performance gains, demonstrating the generalizability of our approach.
Besides, the single-stream CCLM, which achieves better performance by adopting the cross-modal fusion module additionally enhances the interaction between vision and text data. However, this comes at the cost of efficiency, as all possible query-candidate pairs need to be fed into the fusion modules during inference. In contrast, our method CL2CM$^\dag$ is a two-stream architecture, allowing for more efficient calculation of similarity scores. 
Specifically, the training time of our proposed model is approximately 4\% shorter than that of CCLM. In terms of inference time, CCLM takes 11x longer than ours.
\textit{In short, CL2CM$^\dag$ achieves comparable performance with CCLM while achieving a good trade-off between performance and computational cost. These results demonstrate that our method is suitable for large-scale retrieval applications in the real world.}

\subsection{Evaluation on Cross-lingual Video-Text Retrieval} 
The experimental results on VATEX are reported in \cref{tab:sota-vatex}. 
Our CL2CM outperforms all compared methods by a large margin, and these methods all focus on instance-level alignment.          
Moreover, even without resorting to extra pre-training datasets, our CL2CM outperforms the MMP model that was pre-trained on large-scale multi-lingual multi-modal datasets by $4.6\%$ in terms of sumR in text-to-video retrieval.
This result demonstrates the effectiveness of our proposed CL2CM framework.

\subsection{Generalization Analysis}

\begin{table} [tb!]
\renewcommand{\arraystretch}{1.1}
\setlength{\abovecaptionskip}{1.5mm} 
\setlength{\belowcaptionskip}{0cm}
\caption{Zero-shot results on Multi30K. Both methods were pre-trained on MSCOCO.}
\label{tab:sota-zero-shot}
\centering 
\scalebox{1.0}{
\begin{tabular}{@{}l*{12}{r}c @{}}
\toprule
\textbf{Method} & {\textbf{en2de}} & {\textbf{en2fr}}  & {\textbf{en2cs}}\\
\hline
NRCCR \cite{wang2022cross} &{448.7} &{433.8} &{411.2}\\
CL2CM &\textbf{461.2} &\textbf{447.0} &\textbf{428.9} \\
\bottomrule
\end{tabular}
}
\end{table}

To investigate the generalization of our proposed method CL2CM, we compare it with the baseline method NRCCR under a fair comparison setting. Both methods were pre-trained on MSCOCO and evaluated on Multi30K. As shown in [tab:sota-zero-shot], our proposed CL2CM method significantly outperforms NRCCR. We believe that introducing cross-lingual knowledge transfer can help models learn a more comprehensive visual-target language correspondence, thereby improving the generalization ability of the model.

\begin{figure}[tb!]
\centering
\subfigure[Baseline method]{
\label{map-a}
\includegraphics[width=0.46\columnwidth]{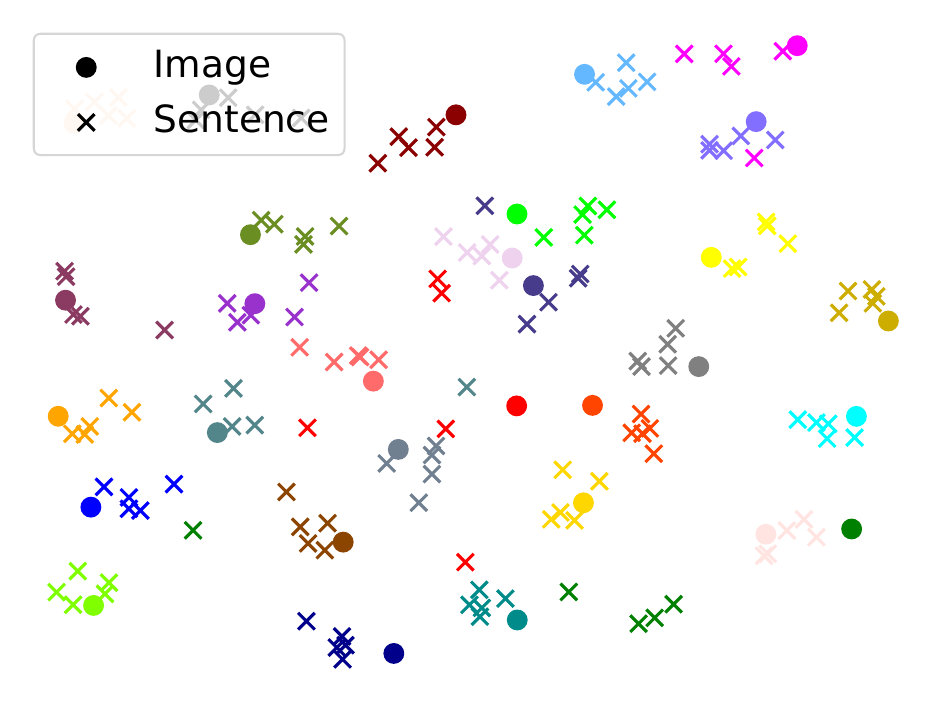}
}
\subfigure[CL2CM]{
\label{map-b}
\includegraphics[width=0.46\columnwidth]{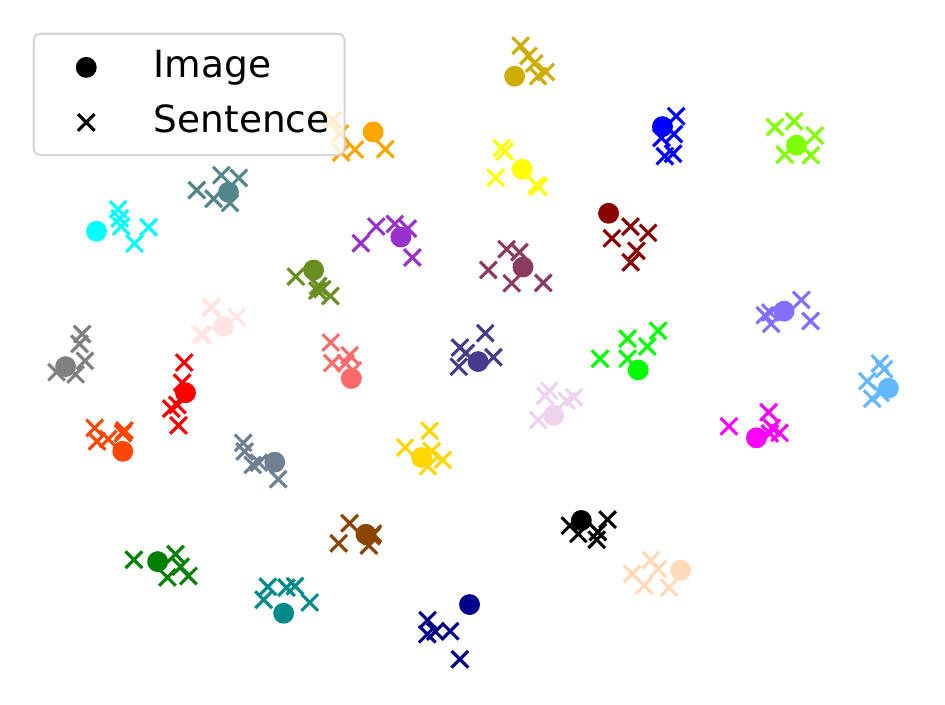}
}
\vspace{-4mm}
\caption{TSNE visualization of 20 images and their corresponding 5 target language (de) sentence representations on Multi30K. Dots with the same color indicate representations belonging to the same class. 
}\label{fig:tsne}
\end{figure}

\subsection{Visualization of Representations}

We visualize image and target-language sentence representations using t-SNE for CL2CM and the Baseline method.
Specifically, we randomly selected 20 images and their corresponding German sentences from the test set of Multi30K, assigning the same color to indicate the same class.
As illustrated in \cref{fig:tsne}, we incorporate the assistance of the CL network built upon the Baseline method, making the image and target language sentence more compact in the vector space.
This visualization result confirms that our CL2CM method effectively improves the alignment quality of image and target language representations.

\begin{figure}[tb!]
\centering\includegraphics[width=1.0\columnwidth]{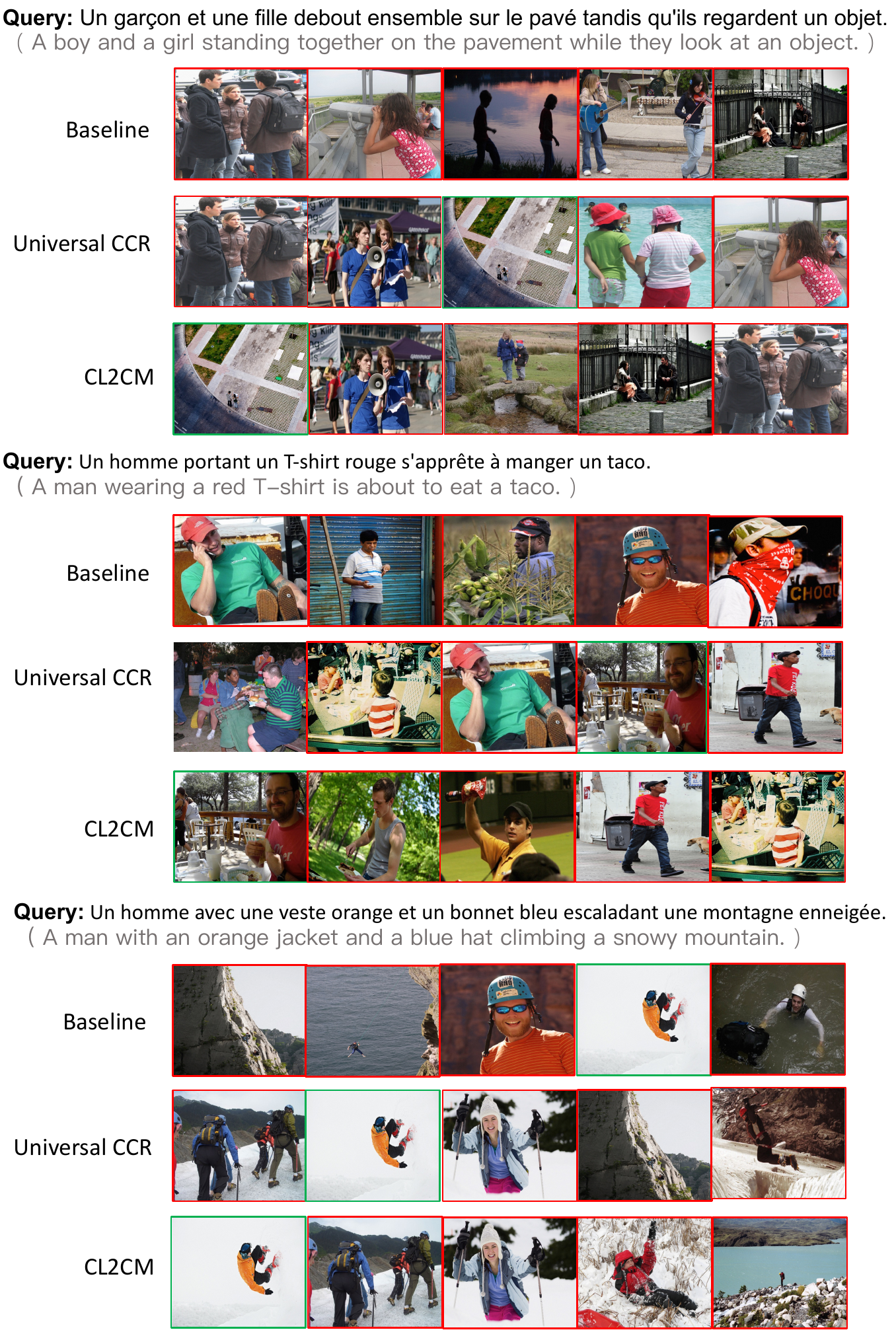}
\vspace{-5mm}
\caption{
The qualitative results of text-to-image retrieval on Multi30K. We compare our CL2CM with the Baseline and Universal CCR methods. 
The first line is text queries in German, 
the next three lines are the top-5 retrieved results of the three different paradigms, respectively.
Correct answers are highlighted in green, while wrong answers are marked in red. The text within the parentheses ($\ $) represents the corresponding English translation. The results show that the top-5 retrieved images not only contain the right objects but also represent the correct relationships among them.
}\label{fig:retrieval_demo}
\vspace{-2mm}
\end{figure}

\subsection{Qualitative Results}
We show some examples of text-to-image retrieval to demonstrate the effectiveness of our proposed CL2CM method in capturing the correspondence between data pairs in different modalities.
As shown in \cref{fig:retrieval_demo}, the top-5 ranking images not only contain the right objects but also accurately represent the semantic relationships among them.
The Baseline method only captures the coarse-grained semantic correspondence between the image and target language, resulting in poorer performance. Furthermore, compared to Universal CCR, the relationship captured by our CL2CM is more comprehensive and accurate, which can be attributed to the coarse-to-fine alignment and relational transfer. 
For instance, the retrieval results in the first example are all semantically related to the objects ``a boy and a girl" and the action ``look at" expressed in the text.
The results demonstrate the superiority of correspondence learning in our proposed CL2CM.

\section{Conclusion} \label{sec:conc}

This paper proposes a new framework, CL2CM, to improve the alignment between vision and target language. By transferring knowledge from the CL network to the CM network, CL2CM is capable of modeling the correspondence between data pairs accurately and alleviating the effect of noisy translations. Extensive experiments on three benchmarks (e.g., Multi30K, MSCOCO, and VATEX) demonstrate the effectiveness of CL2CM and its robustness in the presence of noisy translations.

\begin{small}

\bibliography{main}

\end{small}

\end{document}